\documentclass[runningheads]{llncs}
\usepackage{orcidlink}
\usepackage{graphicx}

\usepackage{booktabs}

\usepackage{pythonhighlight}
\usepackage[frozencache=true,cachedir=minted-cache]{minted} 
\lstnewenvironment{python_code}[1][]{\lstset{style=mypython, frame=none, #1}}{}

\begin{document}
\newcommand{\tagger}{\mathtt{RE\mbox{-}Tagger}}
\title{$\tagger$: A light-weight Real-Estate Image Classifier}
%
\titlerunning{$\tagger$}
\author{Prateek Chhikara \orcidlink{0000-0003-4833-474X} \and
Anil Goyal \and
Chirag Sharma}
\authorrunning{P. Chhikara et al.}
%
\institute{Housing.com, India\\
\email{\{firstname.lastname\}@housing.com}}
\maketitle              
\begin{abstract}
Real-estate image tagging is one of the essential use-cases to save efforts involved in manual annotation and enhance the user experience. This paper proposes an end-to-end pipeline (referred to as $\tagger$) for the real-estate image classification problem. We present a two-stage transfer learning approach using custom InceptionV3 architecture to classify images into different categories (i.e., bedroom, bathroom, kitchen, balcony, hall, and others). Finally, we released the application as \texttt{REST API} hosted as a web application running on 2 cores machine with 2 GB RAM. The demo video is available here\footnote{Demo Video is available at \url{https://www.youtube.com/watch?v=eVWkU7yb-3M}}.
\end{abstract}

\section{Introduction}
\label{sec:intro}
Over the past few years, the demand for online real-estate tools has increased drastically due to the ease of accessibility of the internet, especially in developing countries like India. There are many online real-estate platforms (e.g., Housing.com, Proptiger.com, Makaan.com, etc.) for owners, developers, and real-estate brokers to post properties for buying and renting purposes. Daily, these platforms receive $8,000$ to $9,000$ new listings consisting of approximately $60,000$ to $70,000$ house images belonging to different categories like bedroom, bathroom, kitchen, balcony, living room, etc. To enhance the customer experience, it is necessary to organize the listing images by tagging/categorizing images into one of these categories. Generally, a team of data annotators manually tag a massive volume of images, which is both costly and time-consuming. Moreover, manual tagging introduces a delay of approximately 40 hours from when seller upload the images on the platform to when the listing becomes online.

To overcome these challenges and enhance the user experience, we have developed an end-to-end pipeline for real-estate image tagging (called $\tagger$). For any input image, the $\tagger$ categorizes the image into one of the six categories, i.e., bedroom, balcony, bathroom, kitchen, hall, and others. Concretely, we have used two-stage transfer learning using the custom InceptionV3 \cite{szegedy2016rethinking} architecture for multi-class image classification problem \cite{abou2022transfer}. Finally, we released the pipeline as \texttt{REST API}, which runs in a web browser.
It requires 2 cores machine with 2 GB RAM for hosting the API and can be easily hosted on edge devices.

\section{Model Training and Validation}
\label{sec:approach}
In this section, we present the proposed model architecture along with data acquisition and evaluation results. 

\subsection{Model Architecture}
\begin{figure}[!t]
\centering
  \includegraphics[scale=0.25]{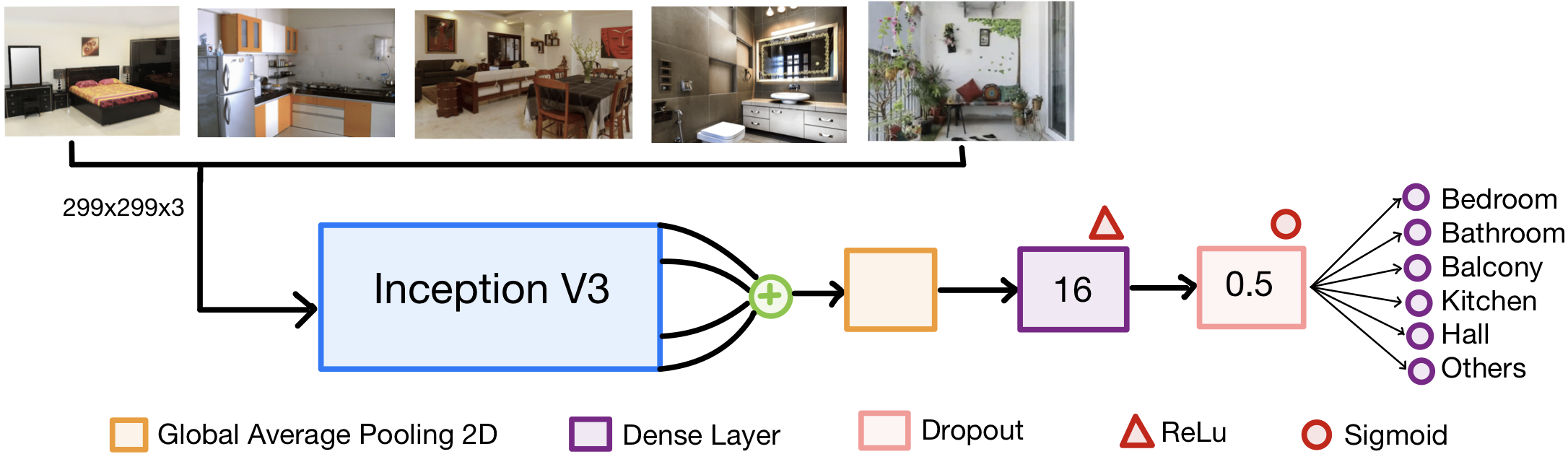}
  \caption{Model Architecture}
  \label{fig:model_architecture}
\end{figure}

We have proposed a two-stage transfer learning approach using the custom InceptionV3 \cite{szegedy2016rethinking} model for the real-estate image classification problem. In the proposed architecture, we have replaced the final classification block of the original InceptionV3 with a global 2D average pooling layer, fully connected layer, dropout layer ($rate=0.5$), and softmax layer. Figure \ref{fig:model_architecture} illustrates the proposed architecture. Please note that, we have experimentally validated that InceptionV3 architecture provides best performance as compared to ResNet \cite{he2015deep}, VGG \cite{simonyan2014very} and Xception \cite{chollet2016xception} architectures on real-estate image classification task. For training the architecture, we have initialized the network with ImageNet weights followed by a two-stage transfer learning approach. 
In the first step, we freeze the base model and only fine-tune the newly added layers (global 2D average pooling, fully connected, dropout, and softmax layers) using Housing.com data. Further, we train the complete end-to-end network on Housing.com data in the second step. We have empirically selected the input image dimensions to be $299\times299\times3$ without cropping and padding. The model training was performed for 50 epochs (both stages) using RMSProp as an optimizer with a learning rate of $0.0001$ and discounting factor ($\rho$) to be $0.9$. We have used categorical cross-entropy as a loss function and set the batch size to 64.

\begin{figure}[!t]
\centering
\includegraphics[scale=0.25]{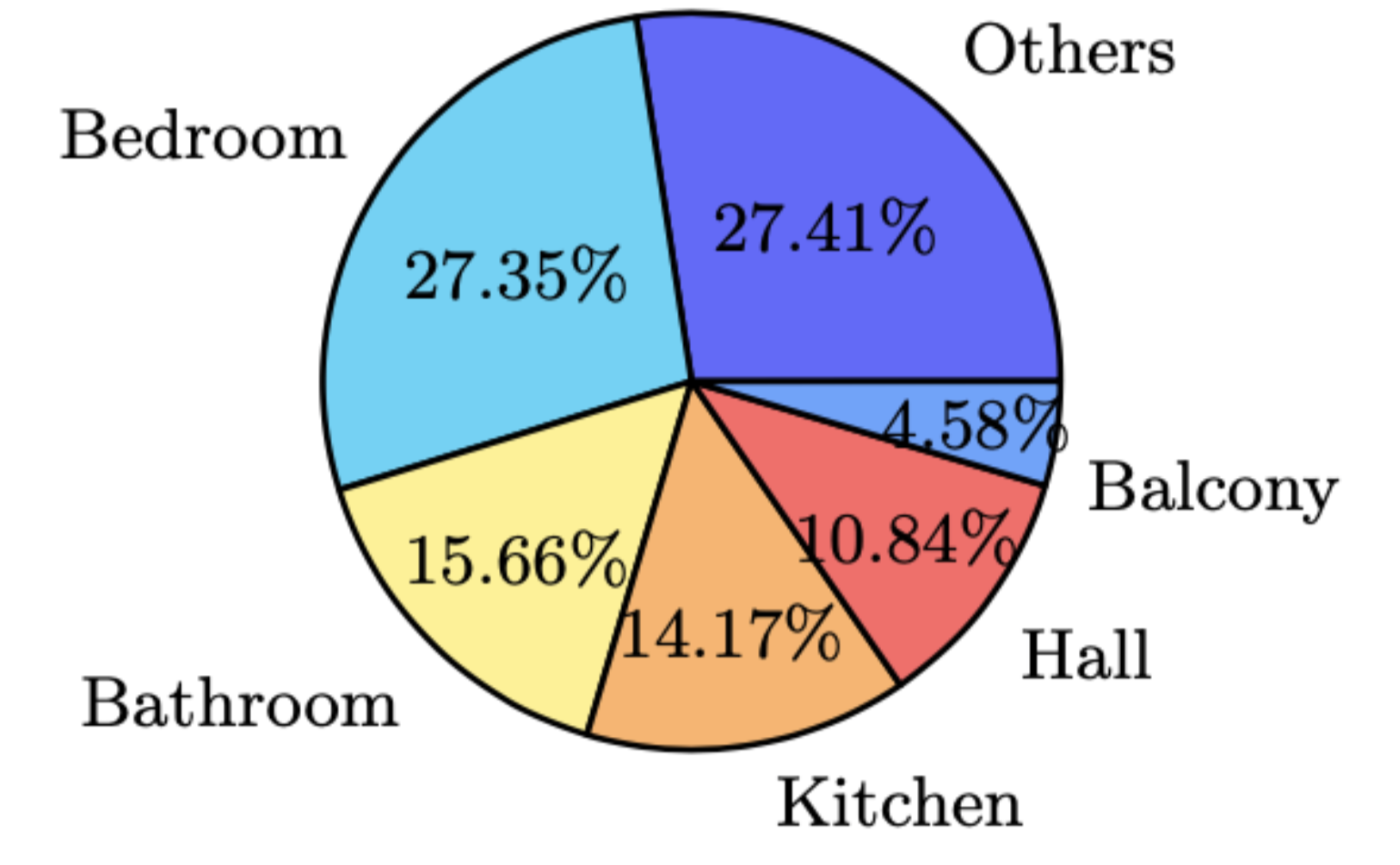}
\caption{The distribution of classes in obtained dataset}
\label{fig:class_distribution}
\end{figure}

\subsection{Data Acquisition}
We have collected $3.1$ million manually annotated images from Housing.com's databases. The majority of examples (approximately $73\%$) in the obtained dataset belong to the bedroom, bathroom, balcony, living, dining, and kitchen classes. 
Moreover, there is a high overlap between dining room and living room classes because residences generally do not have separate living and dining rooms in India. 
Therefore, we considered living and dining rooms a single class, i.e., \textit{`hall'}. The images which do not belong to any of these categories are classified as \textit{`others.'} The detailed distribution of classes is shown in Figure \ref{fig:class_distribution}.

\begin{table}[!t]
\centering
\caption{Obtained precision, recall and F1-scores on the test dataset over all the classes}
\begin{tabular}{lcccccc}
\toprule
\textbf{Class} & 
\textbf{\ Balcony\ }&
\textbf{\ Bathroom\ }  & 
\textbf{\ Bedroom\ }  & 
\textbf{\ Hall\ }  &
\textbf{\ Kitchen\ } & 
\textbf{\ Others\ } \\  \midrule
\textbf{Precision} & 0.98 & 0.98  & 0.87  & 0.84  & 0.85 & 0.82 \\ 
\textbf{Recall}  & 0.82 & 0.98   & 0.89  & 0.94   & 0.95 & 0.98   \\ 
\textbf{F1-score} & 0.90                                   & 0.98  & 0.88  & 0.89 & 0.90 & 0.90  \\ \bottomrule
\end{tabular}
\label{tab:result}
\end{table}

\subsection{Experimental Protocol and Results}
For evaluation, we reserved $100$K images for testing and the remaining for training. For training the model, we randomly under-sample the samples from the majority classes such that all the classes have an equal number of images at the time of training. After under-sampling, we had 1.2 million images consisting of $200$K images from each class. Furthermore, the training dataset is divided into train and validation in the ratio of 9:1. 
Since the classes are imbalanced, we evaluated the learning algorithm in terms of Precision, Recall, and F1-score. Finally, in Table \ref{tab:result}, we present the obtained results over all the classes. The results show that the proposed method performs more than $88\%$ (in terms of F1-score) over all the classes.

\section{REST API and Web Application}
\label{sec:results}
$\tagger$ is developed in \texttt{Python} using Deep Learning frameworks: \texttt{Keras} and \texttt{Tensorflow}. 
We have released the application as \texttt{REST API} which is hosted as a web application running on 2 cores machine with 2 GB RAM. 
Please note that, the API can be easily hosted on edge devices as well.

\begin{figure}[!t]
    \centering
  \fbox{\includegraphics[scale=0.147]{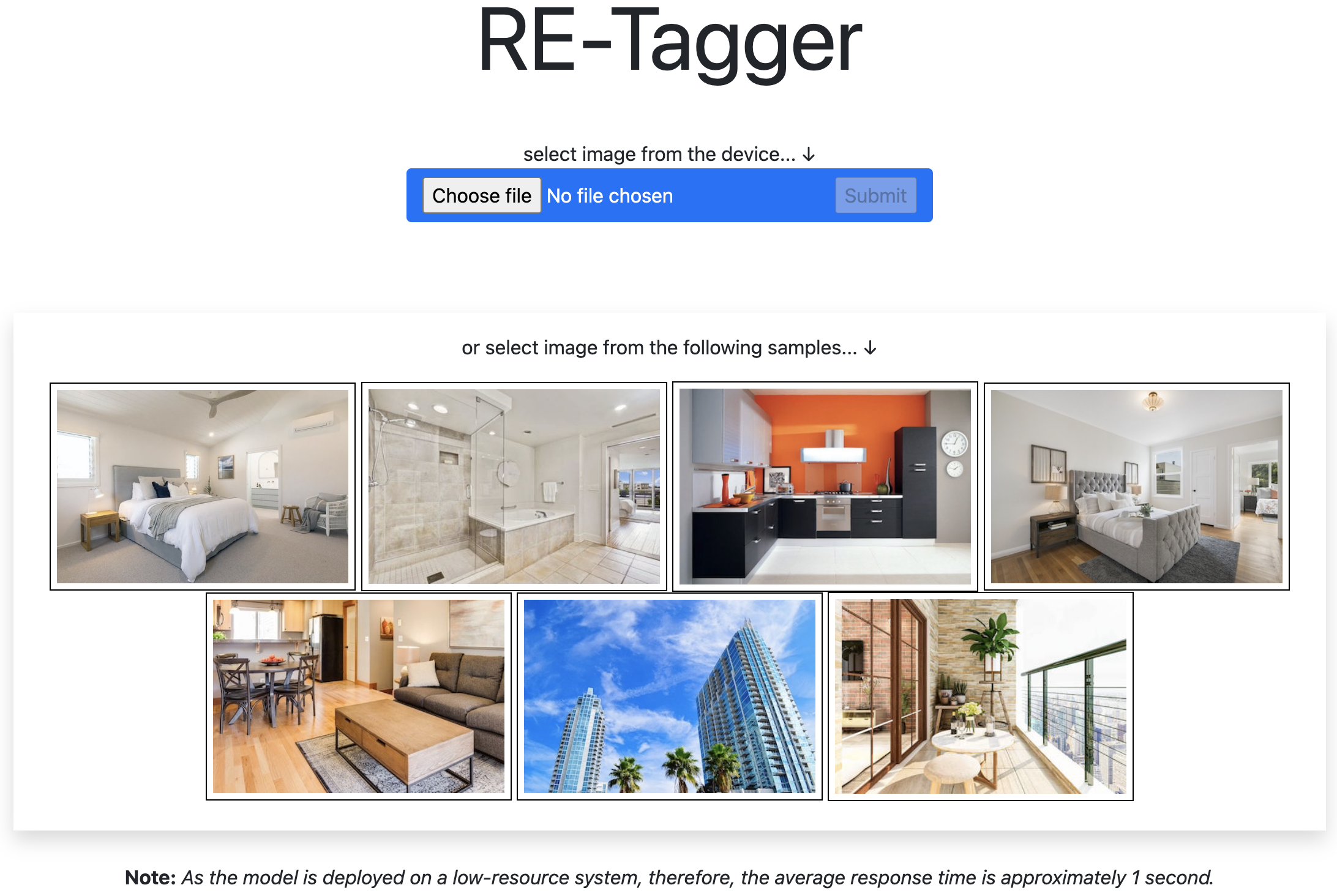}} \ 
  \fbox{\includegraphics[scale=.18,height=.2\textheight]{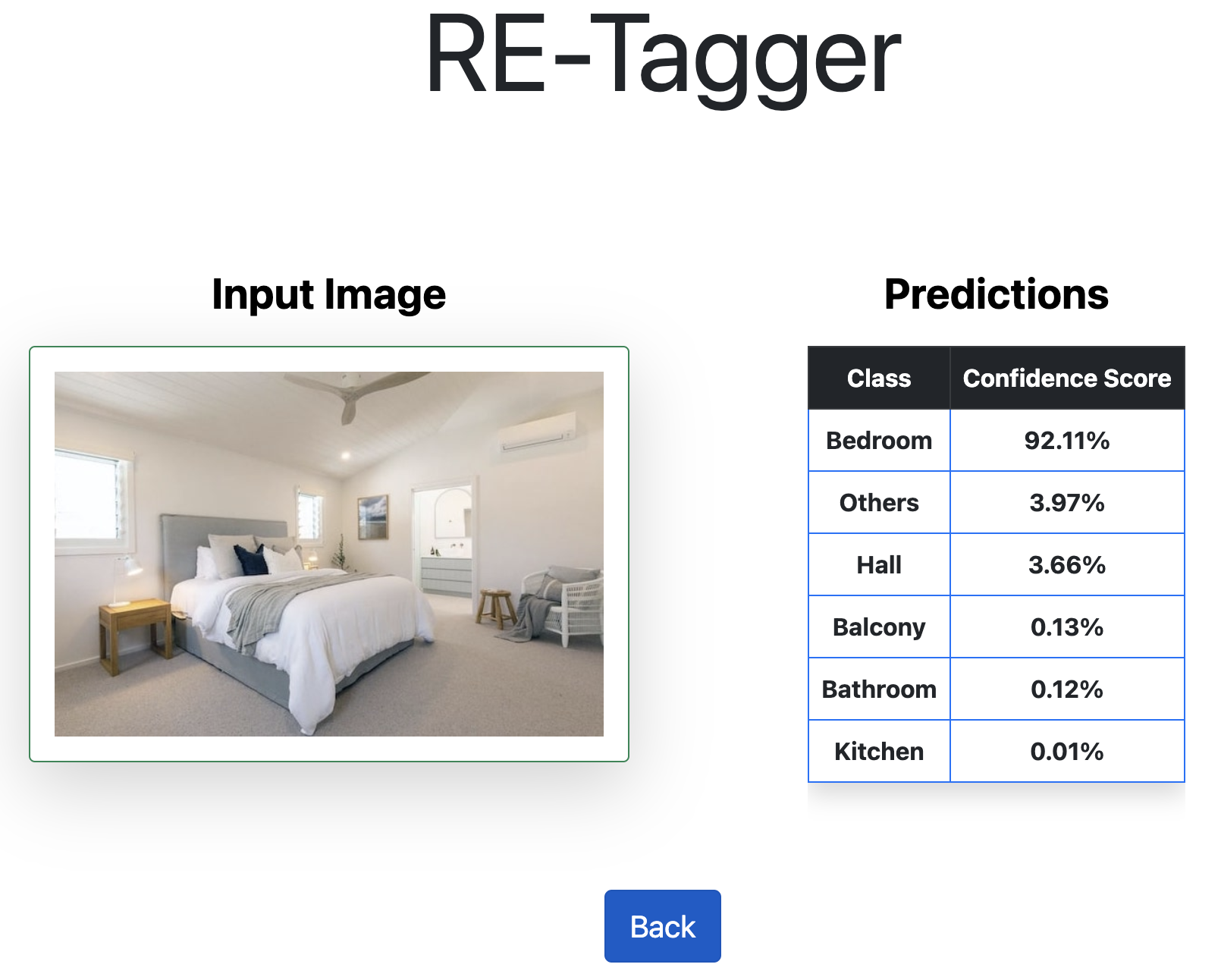}}
  \caption{Web Interface of $\tagger$ API with home page (left) and output page (right)}
  \label{fig:web_interface}
\end{figure}

In Table \ref{tab:code}, we present the \texttt{Python} code snippet along with JSON response for making a HTTP POST request to REST API. The web interface of $\tagger$ is shown in Figure \ref{fig:web_interface} where a user can upload a real-estate image to receive an API response in real-time.

\begin{table}[]
\caption{An example of \texttt{Python} code for making HTTP request to $\tagger$ API url using POST request along with output JSON response}
\begin{tabular}{|c|c|}
\hline
\textbf{Python Code} &  \textbf{JSON Response} \\ \hline
\begin{python_code}
import requests
url='http://127.0.0.1:5000/re-tagger'
filename = 'path_to_file'
files = 
 {'image':(filename, open(filename, 'rb'))}
response=requests.post(url,files=files)
print(response.json())
\end{python_code} 
&
\begin{python_code}
{
"bedroom":"score",
"bathroom":"score",
"balcony":"score",
"kitchen":"score",
"hall":"score",
"others":"score" 
}
\end{python_code}
\\ \hline
\end{tabular}
\label{tab:code}
\end{table}

\section{Conclusion}
 This demo paper introduces the $\tagger$ pipeline that classifies real-estate images into multiple categories: bathroom, bedroom, hall, etc.
We proposed a two-stage transfer learning approach using a custom InceptionV3 model and released the application as REST API hosted as web application.

\bibliographystyle{splncs04}
\bibliography{mybibliography}

\end{document}